\documentclass{article}

\usepackage[final]{corl_2026} 
\usepackage{graphicx}
\usepackage{float}
\usepackage{xurl}
\usepackage{pifont}    
\usepackage{xcolor}    
\usepackage{soul}      
\sethlcolor{yellow}
\usepackage{booktabs}
\usepackage{wrapfig}
\usepackage{amsmath}

\title{PROOF: \underline{P}erceptive \underline{R}obustness via \underline{O}ut-\underline{o}f-Distribution \underline{F}allback for Legged Locomotion}

\author{
    Siyuan Wang$^{1,2}$, Dongjie Zhu$^{3}$, Jinxuan Lu$^{4}$, Wenjun Xu$^{2}$, Jiankun Wang$^{1}$ \\
  $^{1}$Southern University of Science and Technology, $^{2}$Pengcheng Lab, $^{3}$Tsinghua University, \\ $^{4}$The Hong Kong University of Science and Technology (Guangzhou)\\
  \texttt{\small Project Website: \url{https://syw-robotics.github.io/proof-loco/}} \\
}

\begin{document}
\maketitle

\vspace{-2em}
\begin{figure}[H]
    \centering
    \includegraphics[width=0.9\linewidth]{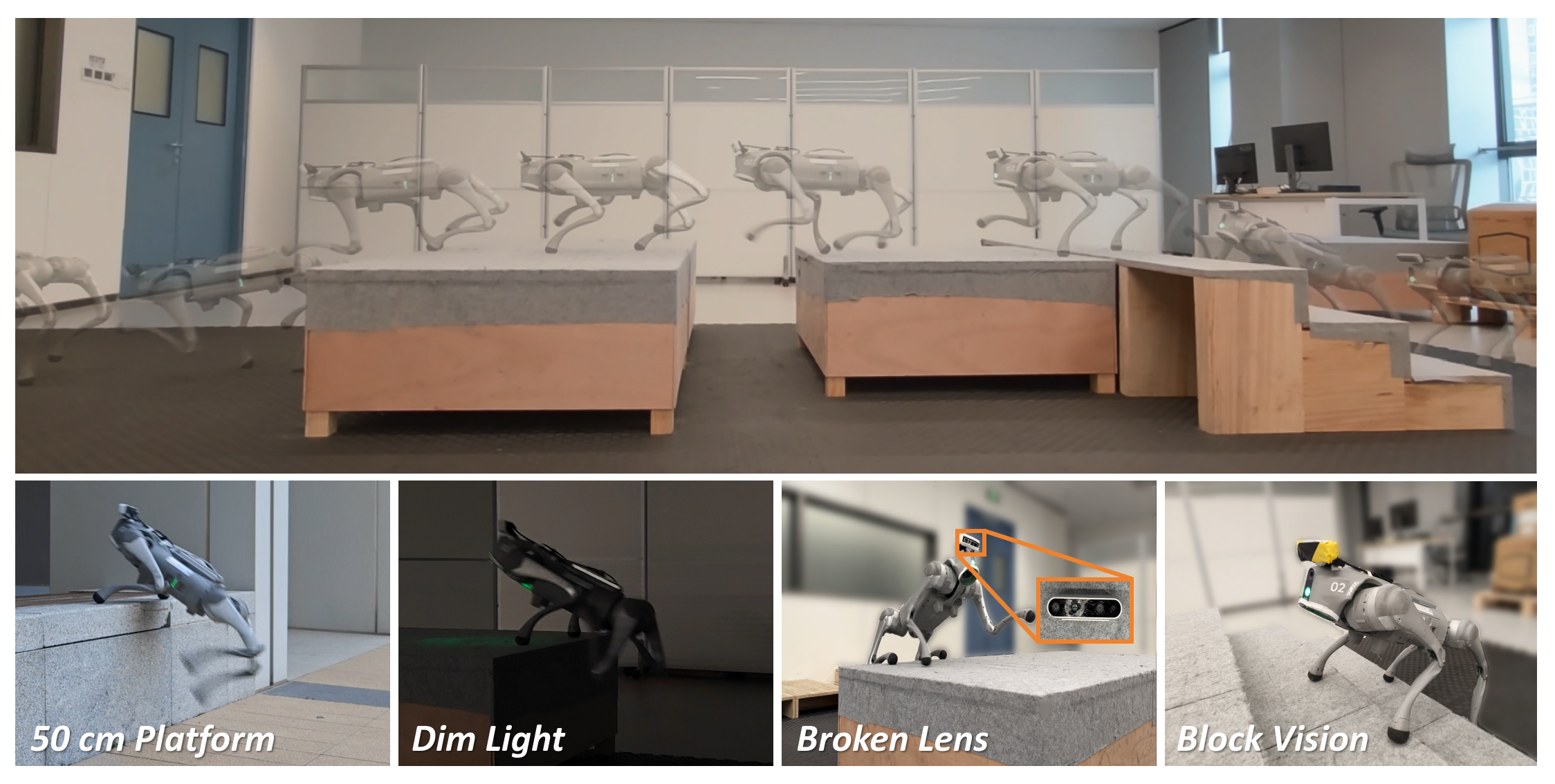}
    \caption{
    \textbf{PROOF} enables agile perceptive robot parkour under challenging real-world conditions.
    The robot traverses large platforms and discontinuous obstacles while remaining reliable under severe visual degradation. When exteroceptive observations become unreliable, \textbf{PROOF} suppresses corrupted visual features and falls back to proprioceptive locomotion behaviours.
    }
    \label{fig:teaser}
\end{figure}

\begin{abstract}

Vision-based perceptive locomotion has enabled legged robots to achieve agile and terrain-aware locomotion across complex environments. However, the reliability of such systems heavily depends on visual observations, which are inherently susceptible to occlusion, reflective surfaces, and illumination variations, often resulting in catastrophic locomotion failures under degraded perception conditions. In this paper, we present \textit{PROOF} (\underline{P}erceptive \underline{R}obustness via \underline{O}ut-\underline{o}f-Distribution \underline{F}allback for Legged Locomotion), a perceptive locomotion framework that enables robust locomotion under severe visual disturbances.
Our method adopts a two-stage teacher–student learning paradigm that explicitly models perception reliability: the teacher learns both perception-driven and proprioceptive fallback behaviors, while the student predicts exteroceptive features with uncertainty and suppresses them under out-of-distribution inputs.
We evaluate our method on a Unitree Go2 quadruped equipped with a RealSense D435 depth camera in real-world settings. Under normal visual conditions, our method enables the robot to successfully perform high-dynamic parkour maneuvers, including traversing large gaps, elevated platforms, and narrow beam terrains. Under severe visual degradation, such as heavy occlusion and strong illumination changes, the robot continues to maintain stable and robust locomotion performance.

\end{abstract}

\keywords{Reinforcement Learning, Perceptive Locomotion}

\section{Introduction}
\label{sec:intro}


Perceptive locomotion has significantly advanced the mobility of legged robots in complex environments \cite{yang2022learning, miki_learning_2022, agarwal_egovision_2023}. By incorporating exteroceptive sensors such as depth cameras and LiDAR, legged robots can proactively perceive surrounding terrain geometry, enabling highly dynamic and agile behaviors including stair climbing, gap traversal, and parkour maneuvers \cite{cheng_extreme_2023, zhuang2024humanoid, hoeller2024anymal, rudin_parkour_2025, lai_wmp_2024, php_parkour_2026, zhuang_deep_2026}. Compared to traditional proprioceptive locomotion policies, perceptive locomotion provides substantially improved terrain awareness and maneuverability. However, the performance of such systems heavily depends on the reliability and accuracy of exteroceptive observations, making perception both the key enabler of agile locomotion and a major source of system vulnerability.

In real-world environments, visual perception is highly susceptible to severe environmental disturbances like sensor occlusion, illumination overexposure, reflective surfaces, motion blur and so on. Such disturbances introduce substantial out-of-distribution (OOD) shifts in visual inputs. For perceptive end-to-end locomotion policies that fully rely on exteroceptive information, corrupted visual observation directly misleads the policy's action , ultimately resulting in unstable locomotion, collisions, or catastrophic failures.

Existing works have attempted to improve the robustness of perceptive locomotion through domain randomization, sensor dropout, adversarial perturbation training, and privileged learning mechanisms\cite{miki_learning_2022, ben2025gallantvoxelgridbasedhumanoid, sun2025learning}. While these approaches improve robustness against perceptual noise and mild distribution shifts, they largely assume that exteroceptive observations remain at least partially informative. Under severe perception degradation, however, visual inputs may become fundamentally unreliable, and continuously relying on corrupted exteroceptive features can further amplify system failure. Therefore, the core challenge is not merely improving robustness to noisy perception, but enabling graceful degradation under catastrophic perception failures. 

To address this challenge, we propose \textit{PROOF} (\underline{P}erceptive \underline{R}obustness via \underline{O}ut-\underline{O}f-distribution \underline{F}allback), a novel two-stage perceptive locomotion framework that efficiently learns agile parkour skills while maintaining resilience against severe visual perception disturbances. To improve robustness against unreliable visual observations, we introduce a two-stage teacher–student framework with uncertainty-aware exteroceptive gating. The privileged teacher is trained with terrain elevation map under random exteroceptive masking, enabling it to learn both perception-driven agile behaviors and proprioception-based fallback locomotion. The resulting behavior is distilled into a depth-vision student, where a recurrent estimator predicts latent exteroceptive features and their uncertainty. This uncertainty serves as an input of an adaptive gate that suppresses corrupted visual inputs and falls back to proprioceptive control under OOD perception. Finally, we validate our method on a real-world Unitree Go2 equipped with a single front-facing RealSense D435 camera, demonstrating agile parkour maneuvers under clean vision and stable locomotion under severe visual degradation.

The contributions of this work are summarized as follows:

\begin{itemize}
\item We present \textit{PROOF}, a perceptive locomotion framework that learns agile quadruped parkour under reliable vision and falls back to proprioception when depth becomes unreliable.
\item We introduce a dual-guard fallback mechanism that detects OOD visual disturbances via latent uncertainty estimation and switches to proprioceptive control when vision becomes unreliable.
\item We validate \textit{PROOF} on a real-world Unitree Go2 hardware with a single RealSense D435, showing robust obstacle traversal and failure avoidance under extreme visual corruptions.
\end{itemize}

\vspace{3em}

\section{Related Work}
\subsection{Learning-Based Legged Locomotion}
Deep reinforcement learning has emerged as the leading paradigm for legged locomotion, allowing quadruped robots to learn motor skills in simulation and then deploy the trained policies onto physical robotic hardware \cite{tan_sim2real_2018,hwangbo_learning_2019,rudin_learning_2022}. Compared to traditional model-based control pipelines, reinforcement learning demonstrates remarkable flexibility and robustness across diverse locomotion scenarios, enabling legged system works across various scenarios like agile locomotion, navigation, force adaptive interaction, motion tracking, etc.\cite{kumar_rma_2021,  hwangbo_deformable_2023, chane2024soloparkour, margolis2024rapid, he2024abs, liao2025beyondmimicmotiontrackingversatile, xu_facet_2025, zhi_learning_2025, zhuang_deep_2026, php_parkour_2026}. 

\subsection{Perceptive Legged Locomotion}
To enable legged robots to traverse unknown and dynamic environments, a large body of work focuses on proprioceptive locomotion, where policies infer terrain properties and robot dynamics from onboard available proprioception without explicit perception input \cite{lee_learning_2020,kumar_rma_2021,kaist_concurrent_2022,nahrendra_dreamwaq_2023,hwangbo_deformable_2023, wang_cts_2024}. These methods achieve impressive robustness on most uneven terrains, but they only react after terrain changes are reflected in contact or body motion, which limits their ability to prepare for obstacles that require anticipatory foothold or body adjustment.

Perceptive locomotion addresses this limitation by incorporating exteroceptive observations such as depth images, reconstructed elevation map and so on. \cite{miki_learning_2022} learns an attention-based recurrent controller that fuses proprioception and exteroception for robust outdoor locomotion. Subsequent works improve visual policy learning through end-to-end cross-modal architectures \cite{yang2022learning, agarwal_egovision_2023, zhuang2023robot, cheng_extreme_2023, zhuang2024humanoid, luo_pie_2024, lai_wmp_2024, hoeller2024anymal, rudin_parkour_2025}. These methods demonstrate that visual perception can substantially expand traversability with lightweight hardware integration, but typically rely on the quality of alignment for visual features between sim and real. This visual sim-to-real gap may be caused by differences in scene richness and imaging mechanisms and unmodeled environmental noise. As a result, corrupted or OOD visual observations directly influence robot's action. Hence, our work focuses on making perceptive locomotion more robust to visual perception, especially for visual degeneration which leads to catastrophic system failure, which is crucial for a safe working robot system.

\section{Method}
\begin{figure}[h]
    \vspace{-2pt}
    \centering
    \hspace*{12pt}%
    \includegraphics[width=\linewidth]{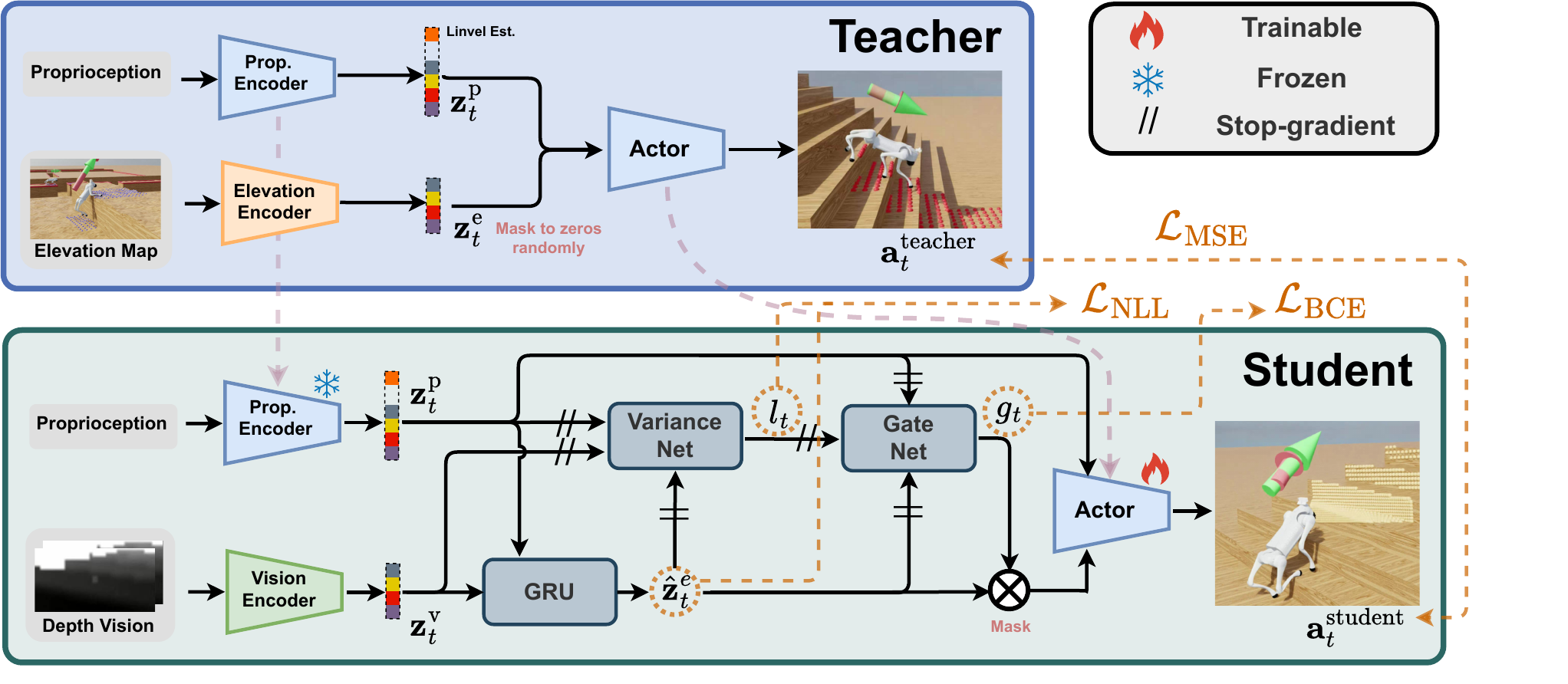}%
    \vspace{-8pt}
    \caption{Our framework consists of two parts: a teacher policy, trained using elevation maps in simulation; and a student policy, trained using depth vision and teacher-student distillation.}
    \label{fig:method-overview}
\end{figure}

As shown in Fig.~\ref{fig:method-overview}, the overall pipeline consists of an elevation-map teacher policy, which is subsequently distilled into a deployable depth-vision student policy. Importantly, we make the teacher and distillation procedure aware of out-of-distribution situations and therefore use a gating mechanism to let the student fallback to safe perception.

\subsection{Parkour as Velocity Tracking}
We formulate agile parkour locomotion as a velocity-tracking problem, where the robot follows commanded planar velocities while interacting with diverse terrain structures. This formulation provides a unified objective across locomotion behaviors, while allowing terrain geometry to implicitly induce complex skills such as gap traversal, beam walking, and platform climbing.


\paragraph{Reward Design}
The teacher is trained with a set of velocity-tracking locomotion rewards consisting of (i) velocity tracking terms, (ii) regularization terms on motion smoothness and energy, and (iii) additional parkour-specific terms. More details are provided in Table~\ref{tab:appendix-rewards} of the Appendix.

Conventional velocity-tracking task reward requires the robot to follow the exactly commanded speed and penalizes deviations from it. During parkour tasks, however, the robot may briefly accelerate beyond the command to build momentum for takeoff, which conflicts with this strict tracking objective. To enable dynamic maneuvers such as jumping over obstacles, we relax the linear velocity tracking objective along the forward ($x$) direction on parkour terrains, and allow the robot to temporarily exceed the target speed to build sufficient momentum for takeoff. The relaxed velocity tracking reward is defined as:
\begin{equation}
    r_v=\exp\!\left(-\frac{(v_x^{\mathrm{cmd}}-\bar{v}_x)^2+(v_y^{\mathrm{cmd}}-v_y)^2}{\sigma^2}\right),\quad
    \bar{v}_x=
    \begin{cases}
        \min(v_x, v_x^{\mathrm{cmd}}+\delta), & \text{parkour terrains},\\
        v_x, & \text{otherwise}.
    \end{cases}
\end{equation}
This modification encourages aggressive yet controlled acceleration before takeoff, while retaining standard tracking behavior on normal terrains.

\subsection{Teacher Policy with Elevation Map-based Perception}

\begin{wrapfigure}{r}{0.45\linewidth}
    \vspace{-8pt}
    \centering
    \includegraphics[width=0.95\linewidth]{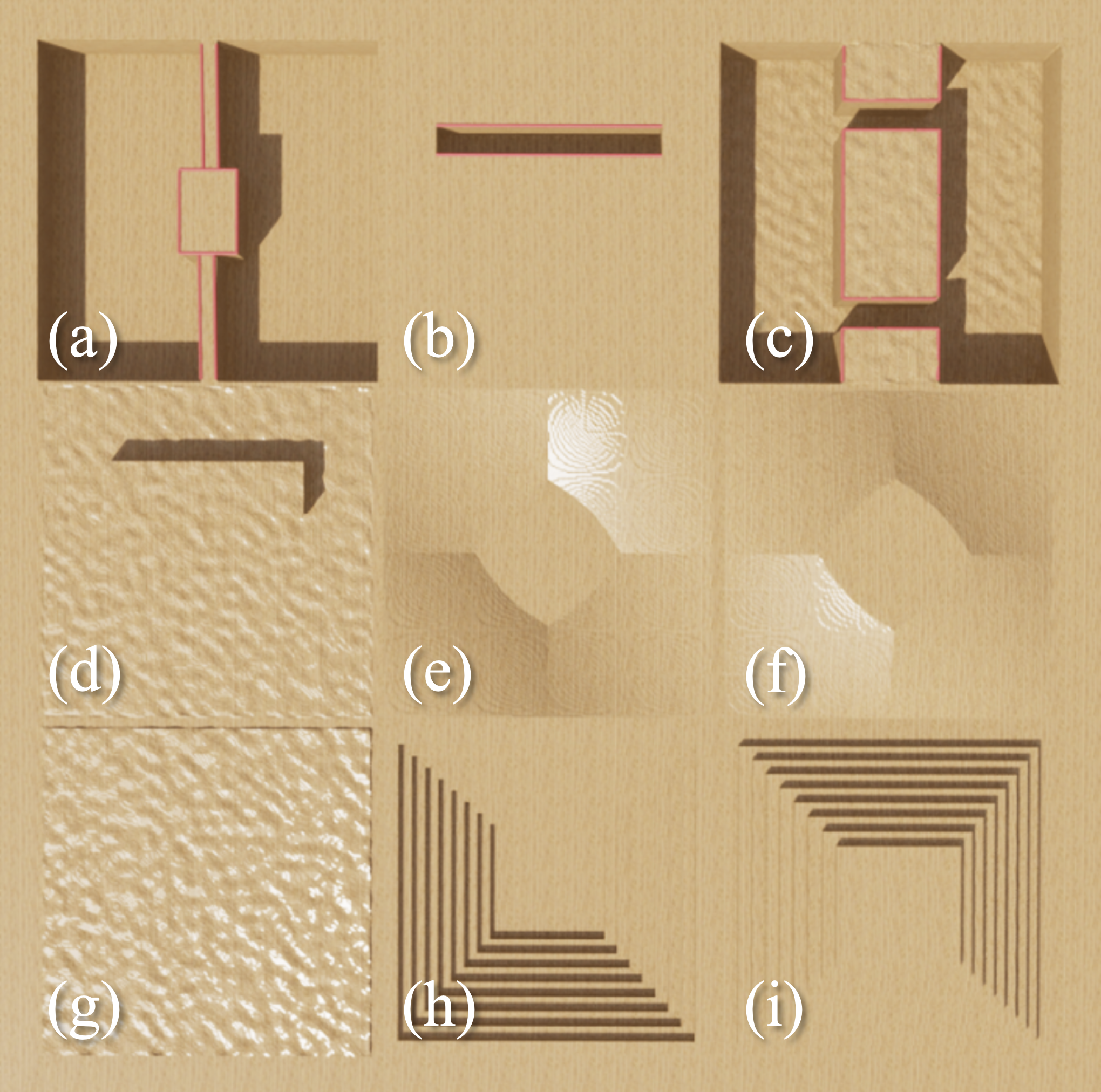}
    \caption{
    Training terrain library used in our experiments. From left to right and top to bottom, the subterrains correspond to: 
    (a) beams, 
    (b) gap, 
    (c) gap with narrow stages, 
    (d) high platform, 
    (e) uphill slope, 
    (f) downhill slope, 
    (g) rough plane, 
    (h) ascending stairs, 
    and (i) descending stairs.
    }
    \label{fig:terrain-types}
    \vspace{-10pt}
\end{wrapfigure}

\paragraph{Teacher Training Design} 
The teacher policy uses elevation maps as exteroceptive observations, which provide accurate local terrain description while remaining low-cost to simulate. 
We adopt an asymmetric actor-critic framework \cite{pinto2018asymmetric} to train the teacher policy. The actor operates at 50 Hz and adopts a \emph{two-branch} architecture. The proprioceptive branch encodes a 10-frame history $o_t^p$ into $z_t^p=E_p(o_t^p)$, where $o_t^p$ consists of base angular velocity, projected gravity, velocity command, joint positions and velocities, and previous actions. The exteroceptive branch encodes the current local elevation map $h_t$ into $z_t^e=E_e(h_t)$. The resulting latent features are used by the policy to generate locomotion actions. While the critic is provided with task-related privileged observations, including base linear and angular velocities, projected gravity, velocity command, joint positions, velocities, efforts, and accelerations, previous actions, foot contact flags, foot heights, and the elevation map. The actions are treated as target joint angles, which are then converted into torque signals by a high-frequency low-level PD controller. In simulation, we adopt a delayed implicit actuator model in Isaac Lab.

Since base linear velocity is critical for velocity tracking but unavailable on the real robot, the first three dimensions of $z_t^p$ are supervised to estimate it \cite{kaist_concurrent_2022}. The action is $a_t^T=\pi_T(z_t^p, m_t z_t^e)$, where $m_t$ masks the exteroceptive latent. The teacher policy is updated by Proximal Policy Optimization (PPO) \cite{ppo}:

\begin{equation}
    \mathcal{L}_T=\mathcal{L}_{\mathrm{PPO}}+\lambda_v \mathrm{MSE}(\hat{v}, v)
\end{equation}


\begin{wrapfigure}{r}{0.30\linewidth}
    \vspace{-8pt}
    \centering
    \includegraphics[width=\linewidth]{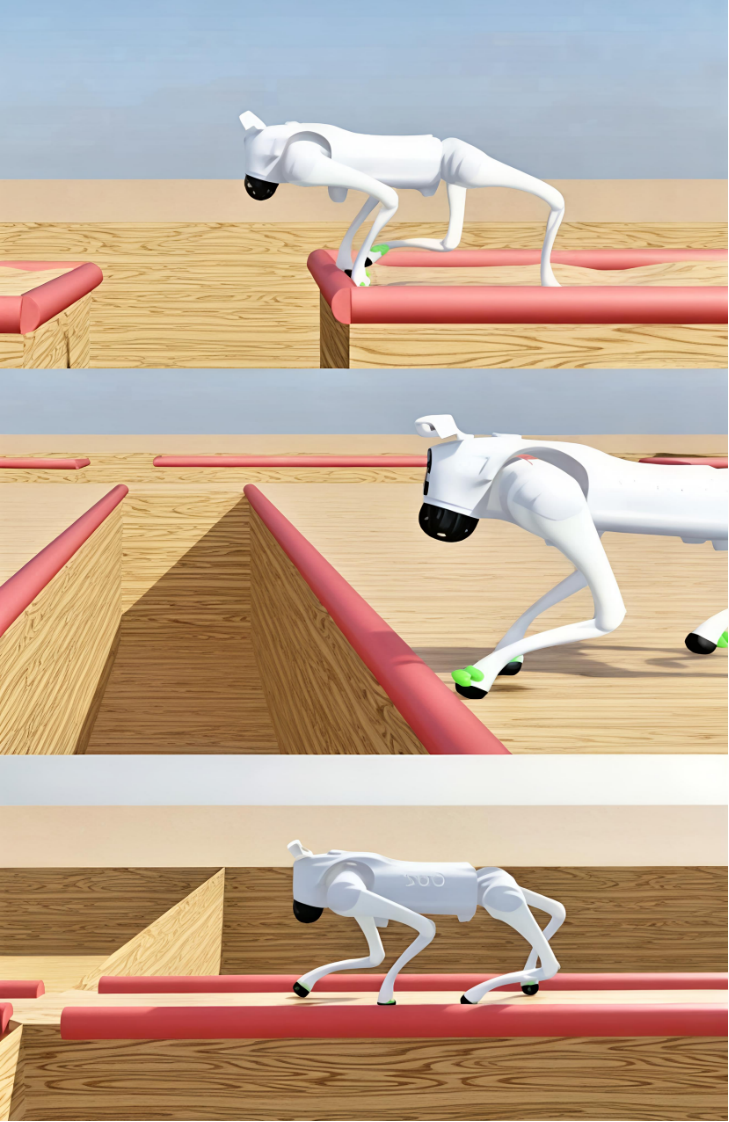}
    \vspace{-10pt}
    \caption{Virtual edge penalty illustration. Red cylinders denote auto-generated virtual obstacles, and green points attached to each foot tip denote virtual monitoring points.}
    \label{fig:edge-penalty}
    \vspace{-20pt}
\end{wrapfigure}

To discourage unstable foot-ground interactions near terrain edges, we detect sharp edges on gap and beam terrains and instantiate virtual cylindrical obstacles around them, similar to the virtual-obstacle design in \cite{hiking_2026}. Four monitoring points are attached to each foot (Fig.~\ref{fig:edge-penalty}). For each point $i\in\mathcal{P}$, let $\mathbf{d}_i$ be its penetration offset into an edge obstacle and $\mathbf{v}_i$ be its linear velocity in the world frame. Following \cite{zhuang2023robot}, we define the edge penalty as
\begin{equation}
    r_{\mathrm{edge}}
    = - \sum_{i\in\mathcal{P}}
    \|\mathbf{d}_i\|\ \cdot\|\mathbf{v}_i\|,
\end{equation}
This term assigns larger penalties to high-speed impacts and scraping motions near obstacle edges, encouraging the policy to select safer footholds.

\subsection{Student Policy with Depth Camera and Uncertainty Fallback}

\paragraph{Student Training with Out-of-Distribution Awareness}
We distill the elevation-map teacher into a deployable depth-vision student using online DAgger-style supervision \cite{pmlr-v15-ross11a}, as shown in Fig.~\ref{fig:method-overview}. The student keeps the teacher's frozen proprioceptive encoder and learns to replace the teacher's elevation-map latent with a depth-based estimate. Concretely, a CNN encodes two delayed 10 Hz depth frames into $z_t^v$, and a GRU predicts the exteroceptive latent mean $\hat{z}_t^e$ from $[z_t^p, z_t^v]$.

\textbf{Uncertainty as OOD signal.}
To assess the reliability of the predicted exteroceptive features, the student models the latent $z_t^e$ as a Gaussian distribution with mean $\hat{z}_t^e$ and diagonal log-variance $\ell_t$. A separate head predicts log-variance $\ell_t$ from $\operatorname{sg}([\hat{z}_t^e, z_t^v, z_t^p])$, where $\operatorname{sg}[\cdot]$ denotes stop-gradient. The model is trained with a negative log-likelihood (NLL) objective:
\begin{equation}
    \mathcal{L}_{\mathrm{NLL}}
    = \frac{1}{2D}\sum_{i=1}^{D}
    \left((z_{t,i}^{e,T}-\hat{z}_{t,i}^e)^2\exp(-\ell_{t,i})+\ell_{t,i}\right).
\end{equation}
This objective encourages \emph{input-dependent uncertainty}: when the depth observation is informative and consistent with the teacher latent, the model minimizes error with small variance; when the observation is ambiguous or corrupted (e.g., under OOD visual disturbances), the model can reduce penalty by increasing $\ell_t$. As a result, the predicted log-variance serves as a learned proxy for \emph{epistemic unreliability of exteroception}, capturing both explicit corruption and intrinsic perceptual ambiguity.

The student is trained jointly with action imitation and latent regression:
\begin{equation}
    \mathcal{L}_S
    = \lambda_a \mathrm{MSE}(a_t^S,a_t^T)
    + \lambda_n \mathcal{L}_{\mathrm{NLL}}(\hat{z}_t^e,\ell_t,z_t^{e,T})
    + \lambda_g \operatorname{BCE}(\tilde{g}_t,m_t),
\end{equation}
where the gate term $\tilde{g_t}$ will be introduced in the next paragraph.

\paragraph{Fallback Mechanism}
Our framework constructs a robust fallback behavior through two complementary supervision pathways:

\textbf{(1) Teacher-induced action-level fallback.}
During training, we inject corrupted depth bursts and align them with an exteroceptive mask $m_t$. When $m_t=0$, the teacher is queried with its exteroceptive latent masked, producing a \emph{proprioceptive-only action}. This provides the student with explicit supervision of \emph{how to act without vision}, ensuring that a valid fallback policy is learned at the action level. Simultaneously, the corresponding teacher latent $z_t^{e,T}$ is set to zero, encouraging the student’s vision branch to represent the absence of reliable exteroception.

\textbf{(2) Uncertainty-aware gating for fallback activation.}
Fitting the exteroceptive latent to zero is not by itself a clean fallback signal: non-zero residual values of $\hat{z}_t^e$ may still affect the policy. We therefore introduce an extra gate that explicitly suppresses exteroception. A gate head maps $\operatorname{sg}([z_t^p, \ell_t, \hat{z}_t^e])$ to a raw score $\tilde{g}_t\in(0,1)$, trained by binary cross-entropy $\operatorname{BCE}(\tilde{g}_t,m_t)$ so that the target is $1$ on clean observations and $0$ under labeled corruption. Simply smoothing this score and multiplying it onto the latent would expose the actor to intermediate latent values. During teacher training, the exteroceptive latent is either fully present or exactly zero, the actor never sees values in between. While, a hard mask in $\{0,1\}$ stays consistent, but the binary gate value might flip between the two states and disturb the policy. Appendix Table~\ref{tab:reliability-gate} therefore tests the gate as a binary classifier at a threshold of $0.5$, and finds that the gate behaviour is acceptably stable. Hence we adopt the hard mask with a hysteresis: a closed gate opens only when $\tilde{g}_t\ge 0.8$, and an open gate closes when $\tilde{g}_t<0.5$. The resulting binary mask $g_t\in\{0,1\}$ enters the policy as
\begin{equation}
    a_t^S = \pi_S([z_t^p, g_t \hat{z}_t^e]).
\end{equation}

\subsection{Additional Design Details}

\paragraph{Terrain Settings} As shown in Fig.~\ref{fig:terrain-types}, the training terrains include normal uneven terrains, high platforms, gaps, beams, and other parkour obstacles. We organize each terrain type into 10 difficulty levels and apply a game-inspired terrain curriculum \cite{rudin_learning_2022}. The key terrain ranges are: climb height $0.20$--$0.70$ m, gap width $0.05$--$0.80$ m, stair height $0.02$--$0.22$ m, and slope magnitude $0.00$--$0.45$. At the end of each episode, the robot is moved to a harder level if it traverses the current terrain, moved to an easier level if it fails to make sufficient progress, and randomly reassigned after solving the highest level. We found that if we spawn robots on both challenging parkour and non-parkour terrains, the policy will struggle to learn proper trot gait. Hence we "warm up" the policy learning on non-parkour terrains first for 1500 iterations, then learns on all terrains.

\begin{figure}[t]
    \centering
    \includegraphics[width=0.75\linewidth]{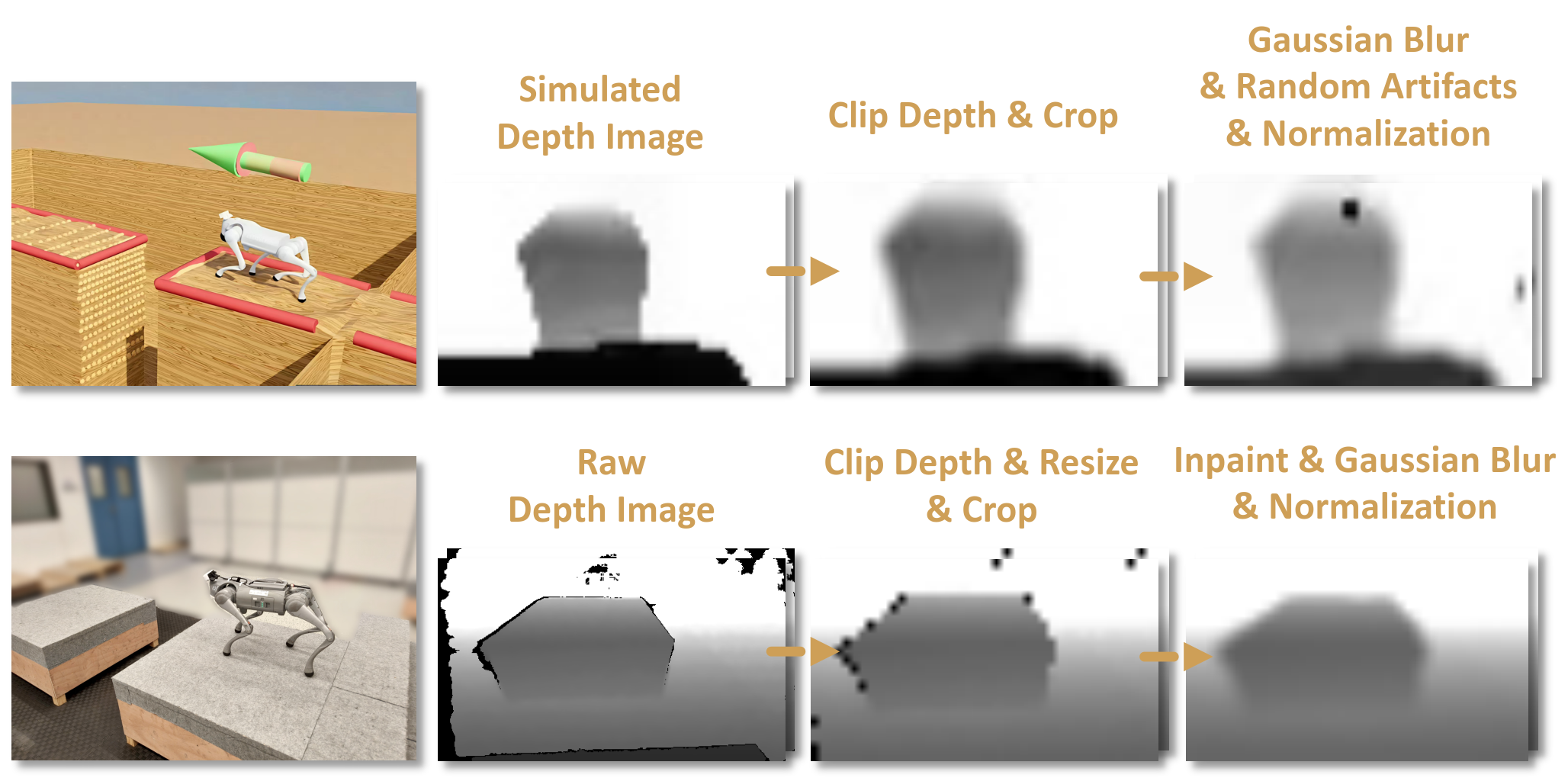}
    \caption{Depth preprocessing used to align simulated ray-caster depth with RealSense observations.}
    \label{fig:depth-preprocess}
    \vspace{-8pt}
\end{figure}

\paragraph{Depth Image Processing}
The student observes the terrain through a forward-facing RealSense D435 depth camera, simulated in Isaac Lab with NVIDIA Warp ray casting and a pinhole camera model. As shown in Fig.~\ref{fig:depth-preprocess}, depth images are clipped to $[0.1,2.0]$ m and preprocessed before being fed to the policy. To reduce the sim-to-real gap, we randomize camera extrinsics, perturb ray origins and directions, and inject minimum- or maximum-depth artifact patches that approximate blind spots and reflective failures. To simulate the depth stream latency, we apply 0.05 s depth frame latency considering the communication cost on real robot hardware. Additional simulation details are provided in the Appendix.

\section{Experiments}

In this section, we evaluate the effectiveness of the proposed framework under corrupted visual observations and real-world parkour agility. Both the teacher and student policies are trained utilizing NVIDIA Isaac Lab~\cite{isaaclab} for high physical fidelity. The teacher and student training take approximately 4.5 hours on a single NVIDIA RTX 4090 GPU to guarantee a robust deployable depth-vision-based policy (teacher for 8000 iterations and student for 4000 iterations). Specifically, we aim to answer the following questions:
1) Does the predicted log-variance reflect visual uncertainty?
2) Is PROOF robust to OOD visual disturbances?
3) Does PROOF achieve strong real-world parkour performance?


\subsection{Behaviour of the learned uncertainty}

\begin{figure}[h]
    \centering
    \begin{minipage}[t]{0.49\linewidth}
        \centering
        \includegraphics[width=\linewidth]{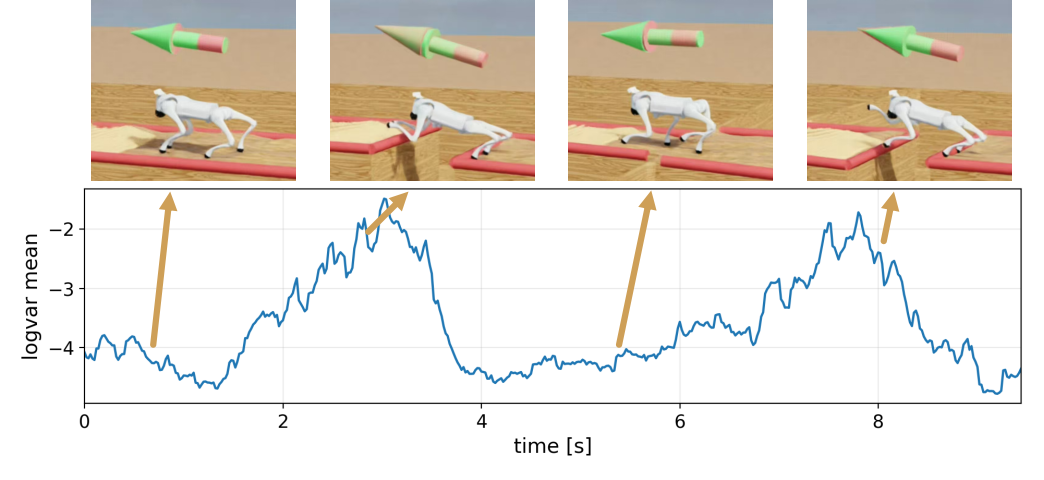}
        \vspace{-6pt}
        \centerline{\small (a) Log-variance during gap traversal.}
    \end{minipage}
    \hfill
    \begin{minipage}[t]{0.49\linewidth}
        \centering
        \includegraphics[width=\linewidth]{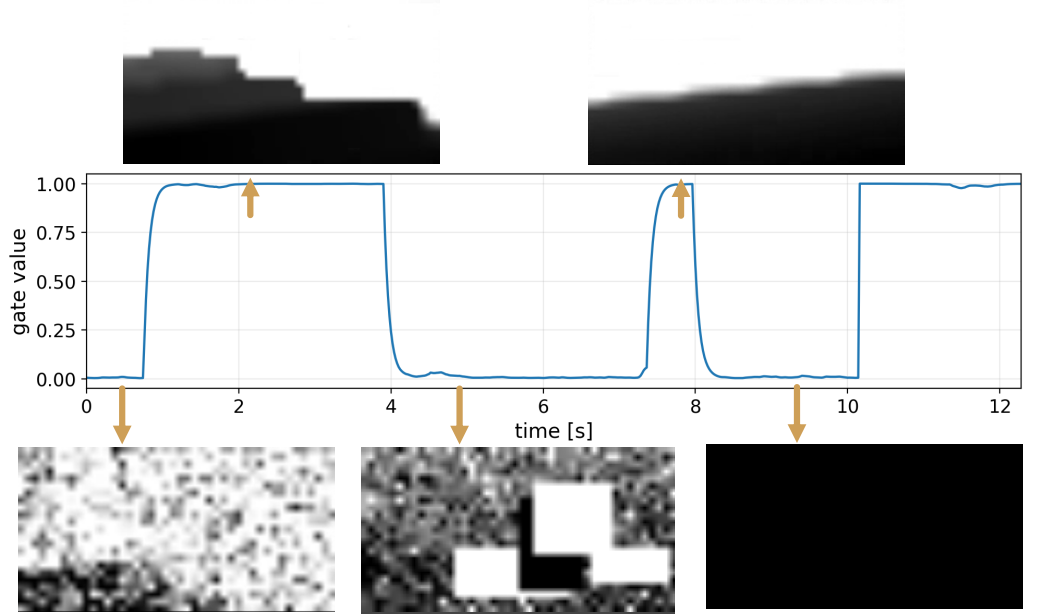}
        \vspace{-6pt}
        \centerline{\small (b) Gate response to depth corruption.}
    \end{minipage}
    \vspace{2em}
    \caption{
    Visualization of the learned uncertainty signal.
    \textbf{(a)} On gap terrains, the predicted mean log-variance remains low before takeoff, rises sharply while the robot is airborne over the gap, and decreases again after landing.
    \textbf{(b)} On stair terrain (as an example of non-parkour terrain), the gate remains open for reliable depth observations and quickly closes when the input is corrupted by strong random noise, large depth artifacts, or full-frame masking.
    }
    \label{fig:learned-uncertainty}
\end{figure}

Fig.~\ref{fig:learned-uncertainty} separates reconstruction uncertainty from the fallback decision.
In the gap example, the depth stream is not externally corrupted, yet log-variance rises during the flight phase because the relevant edge geometry rapidly leaves the camera's view and the exteroceptive latent becomes harder to infer.
In the stair example, severe image degradation causes the gate to close, preventing the corrupted visual features from entering the policy.
These results indicate that the uncertainty head is not merely an auxiliary loss, but provides a reliability signal that can modulate the student policy.

\subsection{Robustness against OOD visual disturbances}

We compare PROOF with two baselines under randomized visual disturbances injected during evaluation. \textbf{Baseline 1} removes exteroceptive masking in teacher training, so the teacher does not learn a proprioceptive fallback behavior. \textbf{Baseline 2} keeps the same teacher masking and corruption schedule, but removes the variance head, the NLL term, and the gate. It regresses the exteroceptive latent with MSE and feeds that latent directly to the actor. Since severe visual disturbances make terrain geometry required for anticipatory parkour unrecoverable, this evaluation measures locomotion on the non-parkour terrains in Fig.~\ref{fig:terrain-types}. All policies are evaluated at terrain level 5, i.e., the middle difficulty level of the curriculum. During rollouts, the injected corruptions are stronger forms of the three identified patterns in training: full-frame depth blocking, large rectangular artifacts, and global random noise (Fig.~\ref{fig:learned-uncertainty}(b)). 

\begin{table}[H]
    \centering
    \caption{
    Simulation robustness under randomized strong visual corruption on non-parkour terrains.
    Each terrain cell reports $R_{v_x} / S$, where $R_{v_x}$ is the mean x-direction velocity tracking reward and $S$ is the traversal success rate.
    A trial is counted as successful if the robot traverses the current terrain segment, following the same terrain evaluation logic as the training curriculum.
    For each method and terrain type, evaluation uses 100 parallel environments and 100 test episodes.
    }
    \label{tab:ood-robustness}
    \vspace{2pt}
    \resizebox{\linewidth}{!}{
    \begin{tabular}{@{}lccccc@{}}
        \toprule
        \textbf{Method}
        & \textbf{Rough Flat}
        & \textbf{Uphill}
        & \textbf{Downhill}
        & \textbf{Upstairs}
        & \textbf{Downstairs}
        \\
        \midrule
        Baseline 1: w/o teacher mask
        & 0.28 / 0 & 0.11 / 0 & 0.35 / 0 & 0.15 / 0 & 0.22 / 0  \\
        Baseline 2: MSE latent distillation
        & 0.83 / 96 & 0.76 / 79 & 0.75 / 84 & 0.69 / 73 & 0.71 / 91 \\
        PROOF
        & \textbf{0.86 / 100} & \textbf{0.79 / 99} & \textbf{0.78 / 100} & \textbf{0.72 / 95} & \textbf{0.93 / 100}  \\
        \bottomrule
    \end{tabular}
    }
    \vspace{-8pt}
\end{table}

Without teacher masking, the student has no reliable fallback target when depth observation becomes out-of-distribution, so Baseline 1 fails under strong visual corruption. Baseline 2 already inherits the masked teacher's fallback behaviour and therefore remains close to PROOF.

Beyond the three corruption patterns seen in training, we further test the gate behaviour on two held-out patterns: randomly oriented stripe masks and reflection-like elliptical artifacts (Appendix~\ref{tab:reliability-gate}). The gate rejects these seen and unseen corruptions while remaining open on clean vision during parkour.

Additionally, the fallback robustness test on hardware under controlled visual corruptions is performed. The designed visual corruptions are physical occlusion, flashlight overexposure, reflection from a waving mirror, coded EMA motion blur, depth dropout, and depth noise, applied on flat ground and on stairs at commanded forward speeds of $0.0$, $0.5$, and $1.0\,\mathrm{m/s}$ (Appendix Table~\ref{tab:realworld-visual-corruption}). Each velocity-terrain pair is tested for five trials. PROOF maintains normal locomotion in nearly all of these conditions without any failure. The remaining failures are strong coded EMA motion blur on stairs at $0.5$ and $1.0\,\mathrm{m/s}$ (success $0.6$ and $0.4$).

\subsection{Real-world parkour evaluation}

Finally, we deploy the policy on a Unitree Go2 with RealSense D435 camera. The ONNX-exported policy runs onboard on an NVIDIA Jetson Nano. We evaluate two representative parkour skills in the real world: gap traversal and high-platform climbing. Although the Go2 reports a maximum joint torque of 45 Nm, using this value in simulation led to a large sim-to-real gap during aggressive motions; reducing the simulated torque limit to 30 Nm substantially improved transfer. Since prior parkour systems use lighter platforms such as Unitree A1, Unitree Go1, and Deep Robotics Lite3 (around 12 kg, while go2 weighs 15 kg) \cite{cheng_extreme_2023, zhuang2023robot, luo_pie_2024}, we do not make direct numerical comparisons with their hardware results.

\begin{figure}[H]
    \centering
    \vspace{-8pt}
    \includegraphics[width=0.8\linewidth]{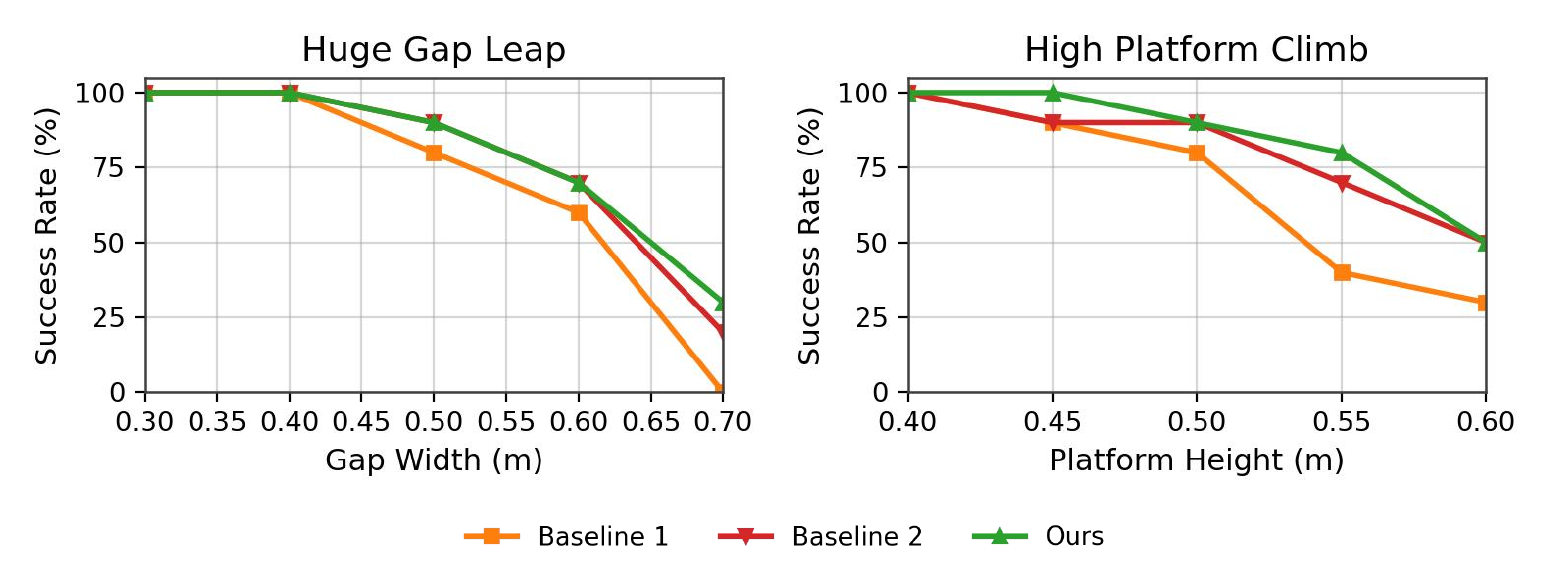}
    \vspace{-8pt}
    \caption{ Real-world parkour deployment on a Unitree Go2.
    We report success rates for two representative high-dynamic skills: leaping increasingly width gaps and climbing increasingly height platforms. }
    
    \label{fig:real-world-parkour}
    \vspace{-6pt}
\end{figure}

As shown in Fig.~\ref{fig:real-world-parkour}, all methods perform reliably at low difficulty, but differs as difficulty increases. PROOF remains competitive on moderate gaps, achieves the highest success rate on the largest tested gap, and maintains strong platform-climbing performance in the mid-to-high difficulty range. These results validate the complete perception-control pipeline under real camera artifacts, latency, calibration errors, and unmodeled contact dynamics.

\section{Conclusion}
\label{sec:conclusion}
We presented \textit{PROOF}, a two-stage perceptive locomotion framework that enables agile quadruped parkour maneuvers and remains robust towards severe visual degradation. The central idea is to treat severe perception failure not as ordinary observation noise, but as a condition under which the controller should stop using exteroception. By training a teacher policy with randomized exteroceptive masking, PROOF learns both perception-guided parkour skills and proprioceptive fallback behaviors.The distilled depth-vision student further models the predicted exteroceptive latent as a Gaussian distribution, allowing the learned log-variance to serve as an interpretable reliability signal for uncertainty-aware gating. Experiments show expressive real-world parkour under clean observations, and stable locomotion when depth perception becomes corrupted or OOD.These results suggest that explicit perception reliability estimation, coupled with a learned fallback policy, is a practical route toward safer deployment of perceptive legged locomotion systems in visually unreliable environments.

\section{Limitations}
\label{sec:limitations}
PROOF is designed to improve robustness under representative severe depth degradation, including occlusion, artifact patches, and strong visual corruption. While these disturbances cover common real-world failure modes, they cannot exhaustively model all possible sensor and environmental conditions. In addition, the predicted log-variance is a relative reconstruction-uncertainty feature for fallback activation, rather than a calibrated measure of terrain-estimation error or a corruption probability.
Future work should reduce the dependence on hand-designed OOD injection and develop more principled uncertainty evaluation protocols.

\acknowledgments{
We thank our co-authors for their valuable contributions. Special thanks go to Botian Xu for his fruitful and constructive discussions, which substantially benefited this work. We also acknowledge Hao Cheng for the camera mount design.

This work was supported in part by the National Natural Science Foundation of China (Grant No. 62506183), the Guangdong S\&T Program (Grant Nos. 2025A0505020002 and 2024B0101010003), the Guangdong Basic and Applied Basic Research Foundation (Grant No. 2025A1515012998), the Guangdong Province Talent Program, the PCL Major Key Project (PCL2025A17-2), the Shenzhen Key Laboratory of Robotics Perception and Intelligence (ZDSYS20200810171800001), and the High-Level Special Funds (G03034K003) from Southern University of Science and Technology, Shenzhen, China.
}

\clearpage

\bibliography{ref} 

\appendix
\makeatletter
\renewcommand{\@seccntformat}[1]{%
  \ifnum\pdfstrcmp{#1}{section}=0
    \appendixname~\csname the#1\endcsname\quad
  \else
    \csname the#1\endcsname\quad
  \fi}
\makeatother


\section{Simulation Setup and MDP Details}

\label{sec:appendix-simulation_mdp}

This section specifies the simulation interfaces and perturbations used by the two-stage training pipeline. Both teacher training and student distillation use $4096$ parallel Isaac Lab environments, $200\,\mathrm{Hz}$ physics, and a $50\,\mathrm{Hz}$ policy step.

\subsection{Observation Space Specification}
\label{sec:appendix-training-obs_spec}

We use separate observation groups for the teacher actor, teacher critic, and deployable student. The teacher actor receives proprioception, a local elevation map, and an exteroceptive latent mask. The critic additionally receives privileged simulation states. During distillation, the student replaces the elevation map with depth images, while the elevation map and mask remain teacher-side supervision.

\begin{table}[H]
\centering
\caption{Observation groups used in teacher training and student distillation.}
\label{tab:appendix-observations}
\small
\begin{tabular}{@{}p{0.18\linewidth}p{0.17\linewidth}p{0.24\linewidth}p{0.31\linewidth}@{}}
\toprule
\textbf{Obs Group} & \textbf{Dim. / Shape} & \textbf{Used by} & \textbf{Contents} \\
\midrule
Proprioception & $10 \times 45$ & Teacher actor, student & Base angular velocity, projected gravity, velocity command, joint positions, joint velocities, and previous actions. \\
Elevation map & $162$ & Teacher actor/critic & Yaw-aligned terrain scan over $1.7\,\mathrm{m}\times0.8\,\mathrm{m}$ with $0.1\,\mathrm{m}$ spacing; ray origin is shifted $0.6\,\mathrm{m}$ forward and heights use a $0.25\,\mathrm{m}$ vertical offset. \\
Depth image$^\dagger$ & $2 \times 23 \times 44$ & Student & Two delayed forward-camera depth frames after ray clipping, crop, artifact noise, blur, and normalization. \\
Exteroceptive mask & $1$ & Teacher actor, distillation loss & Binary gate target; $0$ masks the teacher's exteroceptive latent. \\
\midrule
Privileged state & $80$ & Teacher critic & Base linear/angular velocity, gravity, command, joint positions, velocities, efforts, and accelerations, actions, foot contact flags, and foot heights. \\
\bottomrule
\end{tabular}
\end{table}

\noindent$^\dagger$ The raw depth sensor casts pinhole rays with width $48$, height $27$, FOV $87^\circ\times58^\circ$, and update period $0.05\,\mathrm{s}$. It is mounted on the base at a position of $(0.335,0.015,0.088)\,\mathrm{m}$ with 30 degree downwar in pitch. The sensor returns distance-to-image-plane depth, clips invalid or far rays to the maximum range, crops two pixels from each image border to obtain $44\times23$ (since depth infomations around the frame edges are unreliable for realsense camera), applies random artifacts (at a low probability, this is to simulate the imperfact inpaint on real depth camera) and $3\times3$ Gaussian blur, then clips depth to $[0.1, 2.0]\,\mathrm{m}$ and normalizes it to $[0, 1]$. The observation stacks two frames sampled from the depth history with a $0$--$0.05\,\mathrm{s}$ random delay.

The teacher actor encodes the proprioceptive history and the elevation map into two $64$-dimensional latents. The first three dimensions of the proprioceptive latent are supervised to estimate base linear velocity. The student keeps the same proprioceptive interface and predicts the $64$-dimensional exteroceptive latent from depth through a CNN--GRU estimator.
Noise is applied only to deployable sensing channels: proprioception uses uniform sensor noise, the actor elevation map uses small uniform height noise, and depth images use the vision randomization described below. The privileged state and binary masks are not corrupted.

\subsection{Action Space Specification}
\label{sec:appendix-training-action_spec}

The policy outputs a $12$-D target joint-position offset from the default Go2 posture. Hip joints use scale $0.125$; thigh and calf joints use scale $0.25$.

Targets are executed by a delayed implicit PD actuator solved inside PhysX. We use stiffness $25.0$, damping $0.5$, effort limit $30\,\mathrm{Nm}$, and velocity limit $30\,\mathrm{rad/s}$. At each reset, actuator delay is sampled from $0$ to $3$ physics steps, i.e., $0$--$15\,\mathrm{ms}$.

\subsection{Reward Specification}
\label{sec:appendix-training-reward_spec}

The teacher is trained with a weighted sum of velocity-tracking, regularization, and terrain-safety terms:
\begin{equation}
    r_t=\sum_i w_i r_i.
\end{equation}
The main tracking terms use exponential kernels,
\begin{equation}
    r_{v_{xy}}=\exp\left(-\|e_{v_{xy}}\|_2^2/0.25\right),\qquad
    r_{\omega_z}=\exp\left(-e_{\omega_z}^2/0.25\right),
\end{equation}
where $e_{v_{xy}}$ and $e_{\omega_z}$ are command-tracking errors in the base frame. On parkour terrains, forward overspeed is tolerated by evaluating the tracking error with $\bar{v}_x=\min(v_x,v_x^{\mathrm{cmd}}+0.1)$; otherwise $\bar{v}_x=v_x$.

\begin{table}[t]
\centering
\caption{Reward terms used for teacher training.}
\label{tab:appendix-rewards}
\small
\begin{tabular}{@{}p{0.31\linewidth}p{0.14\linewidth}p{0.45\linewidth}@{}}
\toprule
\textbf{Reward terms} & \textbf{Weight} & \textbf{Function} \\
\midrule
\multicolumn{3}{@{}l}{\textbf{Task rewards}} \\
Linear velocity xy tracking & $1.5$ & $\exp(-((v_x^{\mathrm{cmd}}-\bar{v}_x)^2+(v_y^{\mathrm{cmd}}-v_y)^2)/0.25)$, where $\bar{v}_x=\min(v_x,v_x^{\mathrm{cmd}}+0.1)$ on parkour terrains and $\bar{v}_x=v_x$ otherwise. \\
Angular velocity z tracking & $1.0$ & $\exp(-(\omega_z^{\mathrm{cmd}}-\omega_z)^2/0.25)$. \\
\midrule
\multicolumn{3}{@{}l}{\textbf{Regularization rewards}} \\
Vertical velocity & $-1.0$ & $v_z^2$. \\
Roll/pitch angular velocity & $-0.05$ & $\|\omega_{xy}\|_2^2$. \\
Hip-joint deviation & $-0.2$ & $\|q_{\mathrm{hip}}-q_{\mathrm{hip}}^{0}\|_2^2$, where $q_{\mathrm{hip}}^{0}$ is nomimal joint position. \\
Joint deviation & $-2.5{\times}10^{-3}$ & $\|q-q^0\|_2^2$, where $q^0\|_2^2$ is nominal joint position. \\
Joint acceleration & $-2.5{\times}10^{-7}$ & $\|\ddot{q}\|_2^2$. \\
Joint torque & $-1.0{\times}10^{-4}$ & $\|\tau\|_2^2$. \\
Joint velocity & $-2.0{\times}10^{-4}$ & $\|\dot{q}\|_2^2$. \\
Joint power & $-2.0{\times}10^{-5}$ & $\|\tau\odot\dot{q}\|_1$. \\
Joint-power variance & $-2.0{\times}10^{-6}$ & $\operatorname{Var}(\tau\odot\dot{q})$, this is to discourage overusing certain joint motors. \\
Action rate & $-0.01$ & $\|a_t-a_{t-1}\|_2^2$. \\
Foot air-time & $0.5$ & $\sum_f (t_f^{\mathrm{air}}-0.5)\mathbf{1}_{f,\mathrm{first}}$ under nonzero commands. \\
\midrule
\multicolumn{3}{@{}l}{\textbf{Safety and feasibility rewards}} \\
Alive bonus & $1.0$ & $1$ if the episode is not terminated. \\
Joint power limit & $-10.0$ & $\mathbf{1}\{\max_j |\tau_j\dot{q}_j|>300\,\mathrm{W}\}$. \\
Joint position limit & $-10.0$ & Soft joint-limit violation distance. \\
Foot stumble & $-0.5$ & $\mathbf{1}\{\|F_{xy}^{f}\|>4|F_z^{f}|\}$ for any foot, which is to prevent foot tips from hitting vertical surfaces. \\
Undesired contacts & $-1.0$ & Count of non-foot bodies with contact force above $1\,\mathrm{N}$. \\
Stuck penalty & $-1.0$ & $\mathbf{1}\{|v_x|<0.1, |v_x^{\mathrm{cmd}}|>0.1\}$ plus backward motion under forward command. \\
Terrain Edge penalty & $-1.0$ & $\sum_{i\in\mathcal{P}}\|\mathbf{d}_i\|\|\mathbf{v}_i\|$ for foot monitoring-point penetration into virtual edge obstacles. \\
\bottomrule
\end{tabular}
\end{table}

\subsection{Episode Termination Specification}
\label{sec:appendix-episode_termination_spec}

Episodes are reset either by time truncation or by failure termination:
\begin{equation}
    d_t=d_t^{\mathrm{timeout}}\lor d_t^{\mathrm{ori}}\lor d_t^{\mathrm{contact}}.
\end{equation}
The timeout term is treated as truncation; the other two terms mark failed unfinished rollouts.

\begin{table}[t]
\centering
\caption{Episode termination terms.}
\label{tab:appendix-termination}
\small
\begin{tabular}{@{}p{0.26\linewidth}p{0.18\linewidth}p{0.46\linewidth}@{}}
\toprule
\textbf{Term} & \textbf{Type} & \textbf{Condition} \\
\midrule
Episode length limit & Truncation & Episode length reaches $20\,\mathrm{s}$, i.e., $1000$ policy steps at $50\,\mathrm{Hz}$. \\
Bad base orientation & Failure & $\arccos(-g_{b,z})>1.5\,\mathrm{rad}$, where $g_{b,z}$ is the vertical component of projected gravity in the base frame. \\
Base contact & Failure & The net contact force on robot base link exceeds $5\,\mathrm{N}$. \\
\bottomrule
\end{tabular}
\end{table}

\subsection{Velocity Command Sampling}
\label{sec:appendix-velocity_command_spec}

Velocity commands are resampled every $4$--$8\,\mathrm{s}$. The nominal ranges are
\[
v_x\in[0,1.2]\,\mathrm{m/s},\quad
v_y\in[-0.5,0.5]\,\mathrm{m/s},\quad
\omega_z\in[-1.5,1.5]\,\mathrm{rad/s}.
\]
Small planar commands below $0.2\,\mathrm{m/s}$ are zeroed. For parkour terrains, lateral velocity and heading targets are forced to zero so that the robot approaches obstacles head-on. For beam, crawl, tilt, and stair terrains, the forward command is additionally overridden by $v_x\in[0.8,1.2]\,\mathrm{m/s}$. Other non-parkour terrains use the nominal ranges.

\subsection{Domain Randomization}
\label{sec:appendix-domain_randomization}

\paragraph{Regular Randomization.}
We randomize dynamics properties at startup and inject external perturbations during rollouts. Details are in Table~\ref{tab:appendix-dynamics-dr}.
\begin{table}[t]
\centering
\caption{Regular domain randomization.}
\label{tab:appendix-dynamics-dr}
\small
\begin{tabular}{@{}p{0.34\linewidth}p{0.56\linewidth}@{}}
\toprule
\textbf{D.R. Items} & \textbf{Range / setting} \\
\midrule
Foot friction & Static and dynamic friction in $[0.2,1.5]$; restitution in $[0,0.5]$. \\
Base mass & Additive mass offset in $[-0.5,1.0]\,\mathrm{kg}$. \\
Other body masses & Scale by $[0.8,1.2]$. \\
Base center of mass & Offset each axis by $[-0.05,0.05]\,\mathrm{m}$. \\
Actuator gains & Scale stiffness and damping by $[0.8,1.2]$. \\
Joint friction & Additive friction in $[0,0.1]$. \\
Reset state & Joint positions scaled by $[0.75,1.25]$; base $x$ reset in $[-0.5,0.5]\,\mathrm{m}$. \\
External push & Every $10\,\mathrm{s}$, set base planar velocity in $[-0.5,0.5]\,\mathrm{m/s}$ per axis. \\
\bottomrule
\end{tabular}
\end{table}

\paragraph{Depth Vision Randomization.}
The student depth branch is trained with randomized camera geometry, image corruption, and image frame delay. Details are in Table~\ref{tab:appendix-depth-dr}.
\begin{table}[t]
\centering
\caption{Depth-vision randomization.}
\label{tab:appendix-depth-dr}
\small
\begin{tabular}{@{}p{0.34\linewidth}p{0.56\linewidth}@{}}
\toprule
\textbf{Item} & \textbf{Range / setting} \\
\midrule
Camera translation & Uniform perturbation in $[-0.01,0.01]\,\mathrm{m}$ per axis. \\
Camera rotation & Roll/yaw in $[-2^\circ,2^\circ]$; pitch in $[-8^\circ,8^\circ]$. \\
Ray start point offsets &  Positional in $[-0.005,0.005]\,\mathrm{m}$; Rotational in $[-1^\circ,1^\circ]$ per ray. \\
Sparse artifacts & Random $0/1$ depth artifacts with a probability of 0.0002. \\
Frame delay & $20\,\mathrm{Hz}$ sampling in simulation with $0$--$1$ frame delay, i.e., $0$--$0.05\,\mathrm{s}$. \\
\bottomrule
\end{tabular}
\end{table}

\subsection{Teacher Exteroception Masking}
\label{sec:appendix-visual_disturbance_spec_teacher}

For teacher training, the exteroceptive latent is randomly masked only on non-parkour terrains: flat rough, slope, inverse slope, and $30^\circ$ stairs in both directions. At reset, eligible environments sample a binary mask with probability $0.4$ of closing the exteroceptive branch. The actor therefore sees $\pi_T(z^p,mz^e)$ and must learn both perceptive and proprioceptive fallback behaviors.

\subsection{Student Vision Disturbance}
\label{sec:appendix-visual_disturbance_spec_student}

For student distillation, depth disturbances are enabled on non-parkour terrain envs. A disturbance burst starts with probability $0.01$ per $20\,\mathrm{Hz}$ depth update and lasts $2$--$100$ frames, i.e., $0.1$--$5\,\mathrm{s}$. Each active burst samples one of three modes: (1). \textbf{full-frame blocking} with probability $0.6$ using value $0$ or $1$; (2). \textbf{Gaussian noise} with probability $0.2$ and standard deviation in $[0.15,0.45]$; (3). \textbf{large rectangular artifacts} with probability $0.2$, at most three blocks, $8{\pm}4$ pixels per side, and value $0$ or $1$.

When a depth burst is active, the aligned teacher mask is set to zero, so that student is supervised by both \textbf{teacher's proprioceptive fallback action} and \textbf{closing visual gate target}.

\section{Training Pipeline and Implementation Details}
\label{sec:appendix-training_pipeline}

This section summarizes the execution order of the teacher and student stages. The teacher is first trained with privileged elevation-map perception and randomized exteroceptive masking. The student then distills the frozen teacher into a depth-based recurrent policy with uncertainty-aware gating.

\subsection{Algorithm Overview}
\label{sec:appendix-algorithm_pipeline_overview}

The detailed teacher and student training procedures are summarized in Table~\ref{tab:appendix-teacher-algorithm} and Table~\ref{tab:appendix-student-algorithm}, respectively.

\begin{table}[t]
\centering
\caption{Teacher policy training pipeline.}
\label{tab:appendix-algorithm-flow}
\label{tab:appendix-teacher-algorithm}
\small
\setlength{\tabcolsep}{4pt}
\begin{tabular}{@{}p{0.94\linewidth}@{}}
\toprule
\textbf{Algorithm 1: Teacher Policy Training} \\
\midrule
\textbf{Obs} proprioception $o_t^p$, elevation map $h_t$, privileged state $o_t^{\mathrm{priv}}$, mask $m_t$ \\
\midrule
\texttt{/* Rollout collection */} \\
1. \textbf{for} each rollout step $t$ \textbf{do} \\
2. \quad Sample terrain, command, and domain-randomized dynamics; \\
3. \quad \textbf{if} terrain is non-parkour \textbf{then} sample the exteroceptive mask; \textbf{else} keep perception unmasked; \\
4. \quad Compute $z_t^p\leftarrow E_p(o_t^p)$, $z_t^e\leftarrow E_e(h_t)$, and $a_t^T\leftarrow\pi_T(z_t^p,m_tz_t^e)$; \\
5. \quad Feed the action into the simulator for physics step; \\
6. \quad Compute rewards; \\
7. \quad Store rollout data; \\
\texttt{/* Return and advantage estimation */} \\
8. Compute GAE returns and advantages; \\
9. Normalize advantages over the rollout batch; \\
\texttt{/* PPO update */} \\
10. \textbf{for} each epoch and mini-batch \textbf{do} \\
11. \quad Compute PPO surrogate loss $\mathcal{L}_{\mathrm{clip}}$, value loss $\mathcal{L}_{V}$, and entropy $\mathcal{H}$; \\
12. \quad Compute velocity-estimation supervision $\mathcal{L}_{v}\leftarrow\operatorname{MSE}(\hat{v}_t,v_t)$; \\
13. \quad $\mathcal{L}_T\leftarrow\mathcal{L}_{\mathrm{clip}}+\lambda_V\mathcal{L}_{V}-\lambda_H\mathcal{H}+\lambda_v\mathcal{L}_{v}$; \\
14. \quad Update teacher actor, critic, and encoders with AdamW optimizer; \\
\bottomrule
\end{tabular}
\end{table}

\begin{table}[t]
\centering
\caption{Student distillation pipeline.}
\label{tab:appendix-student-algorithm}
\small
\setlength{\tabcolsep}{4pt}
\begin{tabular}{@{}p{0.94\linewidth}@{}}
\toprule
\textbf{Algorithm 2: Student Distillation} \\
\midrule
\textbf{Obs} proprioception $o_t^p$, depth $d_t$, elevation map $h_t$, mask $m_t$; frozen teacher $\pi_T$ \\
\midrule
\texttt{/* Initialization */} \\
1. Copy the teacher proprioceptive encoder and actor network; \\
2. Freeze the student proprioceptive encoder; \\
\texttt{/* Rollout and online teacher supervision */} \\
3. \textbf{for} each rollout step $t$ \textbf{do} \\
4. \quad Apply depth randomization and burst disturbance; set the teacher's mask per env; \\
5. \quad Compute $\hat{z}_t^e,\ell_t,g_t$ from $(o_t^p,d_t)$ then student action $a_t^S\leftarrow\pi_S(z_t^p,g_t\hat{z}_t^e)$; \\
6. \quad Compute the teacher target $z_t^{e,T}\leftarrow E_e^T(h_t)$ and $a_t^T\leftarrow\pi_T(z_t^p,m_tz_t^{e,T})$; \\
7. \quad Feed the student action into the simulator for physics step; \\
8. \quad Store student outputs, teacher targets, uncertainty, and gate labels; \\
\texttt{/* Distillation update */} \\
9. \textbf{for} each epoch over rollout sequences \textbf{do} \\
10. \quad Recompute student predictions over rollout sequences with recurrent state reset at episode boundaries; \\
11. \quad Compute action-imitation loss $\mathcal{L}_a$, latent MSE $\mathcal{L}_z$, variance NLL $\mathcal{L}_{\mathrm{NLL}}$ on a detached mean, and gate BCE $\mathcal{L}_g$; \\
12. \quad $\mathcal{L}_S\leftarrow\lambda_a\mathcal{L}_a+\lambda_z\mathcal{L}_z+\lambda_n\mathcal{L}_{\mathrm{NLL}}+\lambda_g\mathcal{L}_g$; \\
13. \quad Update $\pi_S$ with AdamW optimizer and gradient clipping; \\
\bottomrule
\end{tabular}
\end{table}

\subsection{Network Architecture and Hyperparameters}
\label{sec:appendix-network_hyperparameters}

The teacher and student network architectures are summarized in Table~\ref{tab:appendix-teacher-network} and Table~\ref{tab:appendix-student-network}, respectively. The teacher actor uses separate proprioceptive and elevation-map encoders, while the teacher critic is a pure MLP that takes privileged state and elevation-map observations. Unless otherwise specified, MLP layers use ELU activations; the depth CNN in the student uses LayerNorm followed by ReLU. Optimization settings are listed in Table~\ref{tab:appendix-training-hyperparameters}. Both stages use $4096$ parallel Isaac Lab environments with AdamW optimization. The full two-stage training takes approximately $4.5$ hours on a single NVIDIA RTX 4090 GPU, with the teacher trained for $8000$ iterations (first 1500 iters are all on non-parkour-terrains) and the student for $4000$ iterations.

\begin{table}[t]
\centering
\caption{Teacher network architecture.}
\label{tab:appendix-teacher-network}
\small
\begin{tabular}{@{}p{0.25\linewidth}p{0.68\linewidth}@{}}
\toprule
\textbf{Module} & \textbf{Specification} \\
\midrule
Proprioceptive encoder & MLP $[256,128]\rightarrow64$; outputs $z_t^p$ from $o_t^p$. The first three latent dimensions are supervised to estimate base linear velocity. \\
Elevation-map encoder & MLP $[128]\rightarrow64$; outputs exteroceptive latent $z_t^e$ from elevation map $h_t$. \\
Mask interface & Actor input is $(z_t^p,m_tz_t^e)$; $m_t=0$ removes elevation-map information and $m_t=1$ keeps it. \\
Actor Network & MLP $[512,256,128]\rightarrow12$; Gaussian policy with initial std. $1.5$. \\
Critic & Pure MLP $[512,256,128]\rightarrow1$; uses privileged state and elevation map. \\
\bottomrule
\end{tabular}
\end{table}

\begin{table}[t]
\centering
\caption{Student network architecture.}
\label{tab:appendix-student-network}
\small
\begin{tabular}{@{}p{0.25\linewidth}p{0.68\linewidth}@{}}
\toprule
\textbf{Module} & \textbf{Specification} \\
\midrule
Proprioceptive branch & Copied from the teacher and frozen during distillation; outputs $z_t^p$. \\
Depth encoder & CNN channels $[4,1]$, $3{\times}3$ kernels, stride $1$, LayerNorm, ReLU; projection MLP $[128]\rightarrow64$ gives visual feature $z_t^v$. \\
GRU estimator & One-layer GRU with hidden size $64$ predicts $\hat{z}_t^e$ from $(z_t^p,z_t^v)$. \\
Uncertainty head & MLP $[64]$ predicts log-variance $\ell_t$ for the latent NLL loss. \\
Gate head & MLP $[64,64]\rightarrow1$ predicts a sigmoid score $\tilde{g}_t$. The actor receives a binary hysteresis mask: a closed gate opens at $0.8$ and an open gate closes below $0.5$. \\
Actor network & Initialized from the teacher; input is $(z_t^p,g_t\hat{z}_t^e)$ and output is a $12$-D Gaussian action. \\
\bottomrule
\end{tabular}
\end{table}

\begin{table}[t]
\centering
\caption{Training hyperparameters.}
\label{tab:appendix-training-hyperparameters}
\small
\begin{tabular}{@{}p{0.24\linewidth}p{0.32\linewidth}p{0.34\linewidth}@{}}
\toprule
\textbf{Item} & \textbf{Teacher PPO} & \textbf{Student distillation} \\
\midrule
\multicolumn{3}{@{}l}{\textit{\textbf{Rollout}}} \\
Parallel envs & $4096$ & $4096$ \\
Horizon & $24$ steps/env & $24$ steps/env \\
\midrule
\multicolumn{3}{@{}l}{\textit{\textbf{Optimization}}} \\
Optimizer & AdamW & AdamW \\
Learning rate & $10^{-3}$ with adaptive KL schedule & $10^{-4}$ \\
Update epochs & $5$ epochs, $4$ mini-batches & $5$ epochs, sequence length $24$ \\
Gradient clipping & Max norm $1.0$ & Max norm $1.0$ \\
\midrule
\multicolumn{3}{@{}l}{\textit{\textbf{Loss and initialization}}} \\
PPO constants & $\gamma=0.99$, $\lambda=0.95$, clip $0.2$, desired KL $0.01$ & -- \\
Loss weights & $\lambda_V=1.0$, $\lambda_H=0.005$, $\lambda_v=1.0$ & action $\lambda_a=2.0$, latent MSE $\lambda_z=2.0$, NLL $\lambda_n=1.0$, gate $\lambda_g=1.0$ \\
Initialization & PyTorch Default Init & Copy teacher proprioceptive encoder and actor network \\
\bottomrule
\end{tabular}
\end{table}

\section{Evaluation Details}
\label{sec:appendix-evaluation_details}

\paragraph{Simulation Evaluation}

In simulation, each policy is evaluated independently on each terrain type with $100$ parallel environments. We collect $100$ completed episodes for every policy--terrain pair. Robots are initialized on the easiest terrain level, and the terrain-level update follows the same curriculum logic as training. Velocity-command sampling and domain randomization are also kept consistent with the training setup. We report the traversal success rate and the mean velocity-tracking reward over the collected episodes.

\paragraph{Real-World Parkour Evaluation}

Real-world parkour evaluations are conducted indoors on the Unitree Go2 with the RealSense D435. Visual corruptions are not injected during these trials. For each policy and terrain setting, we conduct $10$ trials. After each trial, the robot is reset before the next rollout. A trial is successful only if the robot completes the obstacle without falling, triggering an emergency stop, or requiring human intervention.

\subsection{Gate Reliability}
\label{sec:appendix-gate-reliability}

A hard binary gate may switch at high frequency and disturb the policy's action. A continuous gate value in $(0,1)$ is also a unreasonable interface to the actor: the teacher latent is only ever fully present or exactly zero, so intermediate values would have to be absorbed by finetuning the actor during distillation. If the binary gate is reliable, the hard mask is therefore preferred. To verify this, we train with a hard gate and evaluate it as a binary classifier on parkour and non-parkour terrain under seen and unseen corruptions, with gate closure ($g=0$) as the positive class.
AUROC and AUPR use the continuous score. Precision, recall, false-positive rate, and false-negative rate binarize that score at $0.5$. Each condition collects $10{,}000$ policy steps on $100$ parallel environments and is repeated over five random seeds.

On non-parkour terrain, a corruption begins with probability $0.02$ at each camera refresh and lasts $2$--$100$ frames. Seen patterns are those used in training: full-frame masking, global random depth noise, and rectangular depth artifacts. Unseen patterns are randomly oriented stripe masks and reflection-like elliptical artifacts. Clean parkour rollouts report only the false-positive rate, since severe vision corruption leads to inability to parkour. Table~\ref{tab:reliability-gate} shows near-perfect ranking on seen corruptions and strong ranking on held-out patterns. Therefore, it is reasonable to use a final binary gate as a hard mask for fallback behaviour activation.

\begin{table}[H]
\centering
\caption{Gate reliability under seen/unseen non-parkour corruptions and clean parkour vision (mean $\pm$ std)}
\label{tab:reliability-gate}
\small
\setlength{\tabcolsep}{4pt}
\begin{tabular}{@{}lcccccc@{}}
\toprule
\textbf{Condition} & \textbf{AUROC} $\uparrow$ & \textbf{AUPR} $\uparrow$ & \textbf{Prec.} $\uparrow$ & \textbf{Rec.} $\uparrow$ & \textbf{FPR} $\downarrow$ & \textbf{FNR} $\downarrow$ \\
\midrule
Seen, non-parkour & 0.999 & 0.999 & 0.996 & 0.969 & 0.004 & 0.031 \\
Unseen, non-parkour & 0.995 & 0.995 & 0.997 & 0.760 & 0.003 & 0.240 \\
Clean parkour & -- & -- & -- & -- & 0.004 & -- \\
\bottomrule
\end{tabular}
\end{table}

\subsection{Hardware Visual Corruption Tests}
\label{sec:appendix-hardware-corruption}

Table~\ref{tab:realworld-visual-corruption} reports locomotion stablity under controlled visual corruption types. Each corruption is tested for five trials at commanded forward speeds of $0.0$, $0.5$, and $1.0\,\mathrm{m/s}$, on flat ground and on stairs with $30\,\mathrm{cm}$ treads and $15\,\mathrm{cm}$ risers. A trial succeeds if the robot maintains a normal locomotion without failure for the commanded motion. Physical occlusion, flashlight overexposure, and a waving mirror are applied by the operator in the scene. Motion blur is a coded temporal EMA smooth of the depth stream with mixing weight $0.8$. Depth dropout replaces the image by zeros, and depth noise adds Gaussian noise. Depth latency and large calibration shifts are not included: the policy already consumes $10\,\mathrm{Hz}$ depth, which is already tolerant, and small camera-pose error is covered by mild domain randomization.

\begin{table}[H]
\centering
\caption{Hardware success rate under controlled visual corruptions.}
\label{tab:realworld-visual-corruption}
\small
\setlength{\tabcolsep}{5pt}
\begin{tabular}{@{}lccc ccc@{}}
\toprule
& \multicolumn{3}{c}{\textbf{Flat}} & \multicolumn{3}{c}{\textbf{Stairs}} \\
\cmidrule(lr){2-4} \cmidrule(lr){5-7}
\textbf{Corruption} & 0.0 & 0.5 & 1.0 & 0.0 & 0.5 & 1.0 \\
\midrule
Physical occlusion & 1.0 & 1.0 & 1.0 & 1.0 & 1.0 & 1.0 \\
Overexposure (flashlight) & 1.0 & 1.0 & 1.0 & 1.0 & 1.0 & 0.8 \\
Reflection (waving mirror) & 1.0 & 1.0 & 1.0 & 1.0 & 0.8 & 1.0 \\
Motion blur & 1.0 & 1.0 & 1.0 & 1.0 & 0.6 & 0.4 \\
Depth dropout & 1.0 & 1.0 & 1.0 & 1.0 & 1.0 & 1.0 \\
Depth noise & 1.0 & 1.0 & 1.0 & 1.0 & 1.0 & 1.0 \\
\bottomrule
\end{tabular}
\end{table}

\end{document}